\PassOptionsToPackage{usenames}{xcolor}
\PassOptionsToPackage{dvipsnames}{xcolor}

\documentclass[sigconf]{acmart}
\AtBeginDocument{%
\providecommand\BibTeX{{%
\normalfont B\kern-0.5em{\scshape i\kern-0.25em b}\kern-0.8em\TeX}}}

\usepackage{cleveref}

\usepackage{subfig}

\usepackage{enumitem}

\usepackage{booktabs} 

\usepackage{csquotes}
\usepackage{amsmath}
\usepackage{outlines}

\usepackage{eurosym}
\usepackage{siunitx}
\usepackage[usenames]{xcolor}

\usepackage{graphicx}

\usepackage{pifont}
\usepackage{array}
\newcolumntype{P}[1]{>{\centering\arraybackslash}m{#1}}
\newcolumntype{L}[1]{>{\arraybackslash}m{#1}}

\usepackage{xspace}
\newcommand\ie{i.\,e.\xspace}
\newcommand\eg{e.\,g.\xspace}

\newcommand\cf{cf.\xspace}

\usepackage{url}
\usepackage{balance}

\newcommand{\xmark}{\ding{55}}%

\makeatletter
\renewcommand{\fps@figure}{htb}         
\renewcommand{\fps@table}{htb}         
\makeatother

\allowdisplaybreaks

\setcopyright{iw3c2w3}

\begin{document}

\title[Integrating Floor Plans into Hedonic Models for Rent Price Appraisal]{Integrating Floor Plans into Hedonic Models for\\ Rent Price Appraisal}

\author{Kirill Solovev}
\email{kirill.solovev@wirtschaft.uni-giessen.de}
\affiliation{
	\institution{University of Giessen}
	\streetaddress{Licher Str.\ 62}
	\country{Germany}
}

\author{Nicolas Pröllochs}
\email{nicolas.proellochs@wi.jlug.de}
\affiliation{
	\institution{University of Giessen}
	\streetaddress{Licher Str.\ 62}
	\country{Germany}
}

\renewcommand{\shortauthors}{Kirill Solovev and Nicolas Pröllochs}

\begin{abstract}
	Online real estate platforms have become significant marketplaces facilitating users' search for an apartment or a house.
	Yet it remains challenging to accurately appraise a property's value.
	Prior works have primarily studied real estate valuation based on \emph{hedonic price models} that take structured data into account while accompanying unstructured data is typically ignored.
	In this study, we investigate to what extent an automated visual analysis of apartment \emph{floor plans} on online real estate platforms can enhance hedonic rent price appraisal.
	We propose a tailored two-staged deep learning approach to learn price-relevant designs of floor plans from historical price data.
	Subsequently, we integrate the floor plan predictions into hedonic rent price models that account for both structural and locational characteristics of an apartment.
	Our empirical analysis based on a unique dataset of \num{9174} real estate listings suggests that current hedonic models underutilize the available data.
	We find that (1) the visual design of floor plans has significant explanatory power regarding rent prices -- even after controlling for structural and locational apartment characteristics, and (2) harnessing floor plans results in an up to 10.56\% lower out-of-sample prediction error.
	We further find that floor plans yield a particularly high gain in prediction performance for older and smaller apartments.
	Altogether, our empirical findings contribute to the existing research body by establishing the link between the visual design of floor plans and real estate prices.
	Moreover, our approach has important implications for online real estate platforms, which can use our findings to enhance user experience in their real estate listings.
\end{abstract}

\copyrightyear{2021}
\acmYear{2021}
\acmConference[WWW '21]{Proceedings of the Web Conference 2021}{April 19--23,
	2021}{Ljubljana, Slovenia}
\acmBooktitle{Proceedings of the Web Conference 2021 (WWW '21), April 19--23,
	2021,Ljubljana, Slovenia}
\acmPrice{}
\acmDOI{10.1145/3442381.3449967}
\acmISBN{978-1-4503-8312-7/21/04}

\ccsdesc[500]{Applied computing~Electronic commerce}
\ccsdesc[500]{Applied computing~Economics}
\ccsdesc[300]{Information systems~Decision support systems}
\ccsdesc[300]{Computing methodologies~Modeling methodologies}
\ccsdesc[300]{Computing methodologies~Neural networks}

\keywords{Online real estate platforms, hedonic price models, floor plans, visual analytics, image sentiment}

\maketitle

\section{Introduction}


Online real estate platforms have become significant marketplaces for the real estate industry.
These online platforms provide an overview of available properties in a specific location and facilitate users' search for an apartment or a house \citep{Yuan.2013}.
In 2011, one of the biggest online real estate providers, Zillow, reported \num{24} million unique visitors.
Today, this number has risen to \num{196} million and stays on the upwards trend, indicating continual interest from prospective home buyers and tenants \citep{Zumpano.2003}.
While the number of people referring to online real-estate platforms continues to grow, there is a rising need for tools and recommendation algorithms that allow for an accurate appraisal of real estate prices \citep[\eg][]{Yuan.2013}.


Prior works have primarily studied real estate valuation based on \emph{hedonic price models} \citep{Wallace.1997, Monson.2009, Sopranzetti.2010}.
Hedonic pricing theory suggests that the valuation of housing values or rents can be viewed as a weighted sum of its features \citep{Wallace.1997}.
Specifically, the price and rent are determined by the attributes and characteristics of the dwelling unit and the surrounding neighborhood.
Hedonic models provide a high degree of flexibility when selecting the attributes, which can roughly be divided into structural and locational attributes \citep{Natividade.2007}.
The structural factors describe an apartment itself (\eg number of rooms, amenities, parking, pool), while the locational attributes are composed of external features affecting the price.
For instance, previous studies \citep[\eg][]{Peterson.2009, Sirmans.2005} have examined the impact of the distance to transport hubs and shopping centers on real estate prices.
Altogether, the hedonic approach provides scholars and practitioners with a framework of assessing the value of real estate properties, where the final price entails a precise valuation of a diverse feature package.


While structural and locational attributes are of great importance for real estate price models, modern online real estate marketplaces provide additional information that has been neglected in previous works.
A particularly relevant feature for housebuyers and prospective tenants may be real estate \emph{floor plans}.
According to rightmove.co.uk, \SI{90}{\percent} of home-buyers think that a floor plan is an essential part of the decision making process when finding an apartment or house.
A real estate floor plan is a schematic illustration that provides a top-down view of the property.
While information about floor plans may be partially provided in structured form (\eg, in floor plan filings), the actual \emph{floor plan images} may contain price-relevant hidden information.
For example, the relationship between rooms and spaces, as well as the overall layout are typically only available from the floor plan image itself. 
Information on floor plans is typically also not available in textual form as the vast majority of platforms either do not include an alternative text description of the floor plan or the text is very general and not helpful \citep{Goncu.2015}.

Altogether, current hedonic models for rent price appraisal fail at incorporating the complex data in floor plans, as it is not typically a part of the structured information.
Thus, in this study, we investigate to what extent an automated visual analysis of \emph{floor plan images} on online real estate marketplaces can help to enhance real estate appraisal.

\vspace{0.2cm}
\noindent\textsc{Research Question:} \emph{Can apartment floor plans on online real estate platforms enhance hedonic rent price appraisal?}
\vspace{0.2cm}


\vspace{0.2cm}
\textbf{Methodology: }
We propose a tailored two-staged deep learning approach to determine the hedonic price of apartment floor plans on online real estate marketplaces.
Our method does not require any kind of manual labeling by human raters, as it learns price-relevant designs of floor plans based solely on price data.
In the first step, we compute adjusted rent prices by regressing the monthly rent price on the structural and locational variables.
This allows us to extract the variance in rent that is unexplained by these variables and prevents our model from learning information that is already available from known features.
Second, we use convolutional neural networks (CNN) and the adjusted rent prices to predict the sentiment of the floor plans, \ie the apartment rent price after controlling for locational and structural characteristics of an apartment.
We then evaluate the benefits of integrating floor plans into hedonic rent price models as follows: (i)~we perform hedonic regression analysis to evaluate the \emph{explanatory power} of floor plans, and (ii)~we evaluate the out-of-sample \emph{prediction performance} in rent price prediction.


\vspace{0.2cm}
\textbf{Main Findings: }
We collected a unique dataset of \num{9174} real estate listings from a leading online real estate marketplace in Germany.
For each apartment, we collected a raw image showing the floor plan of the apartment, the monthly rent price, the monthly rent price per $\text{m}^2$, and \num{36} fields with structural and locational attributes about the apartment.
Based on a thorough hedonic analysis of rent prices, we yield the following main findings:
\begin{itemize}
	\item The visual design of floor plans has significant explanatory power in hedonic regression models for rent price appraisal -- even after controlling for structural and locational apartment characteristics.
	\item Harnessing floor plans results in an up to \SI{10.56}{\percent} lower out-of-sample prediction error of rent prices.
	\item The predictive power of floor plans varies by the nature of the floor plans.
	      Floor plans yield a particularly high gain in prediction performance for older and smaller apartments.
\end{itemize}

Altogether, our empirical findings contribute to the existing research body by quantifying the hedonic value of floor plans and establishing the link between visual designs of floor plans and rent prices.
We show that there is an underutilization of the available data in current hedonic models.
To the best of our knowledge, our paper is the first study that demonstrates how harnessing floor plans can enhance real estate appraisal on online real estate platforms.

\vspace{0.2cm}
\textbf{Implications: }
Our findings have direct implications for online real estate platforms, providing avenues to improve user experience in their real estate listings.
Our results also allow practitioners in the real estate industry to enhance the accuracy of their price estimates without the need for costly and subjective human annotations.
The latter equips real estate investors with the clear benefit of improved risk assessment, allowing them to enhance their portfolios and reduce the possibility of misvaluation.

\begin{figure*}
	\captionsetup{position=top}
	\centering
	\subfloat[$\boldsymbol{\mathit{Rpms}}$]{\includegraphics[width=.3\linewidth]{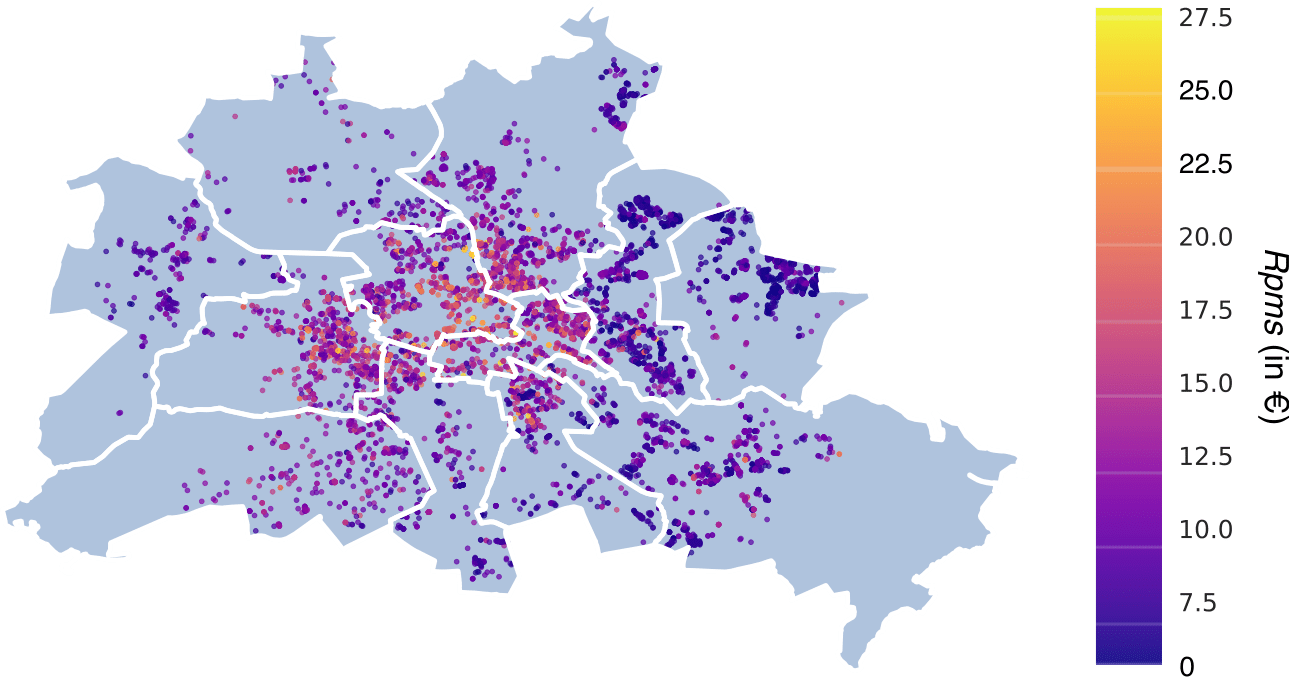}\label{fig:berlin_districts_rpms}}
	\hspace{0.5cm}
	\subfloat[$\boldsymbol{\mathit{Rent}}$]{\includegraphics[width=.3\linewidth]{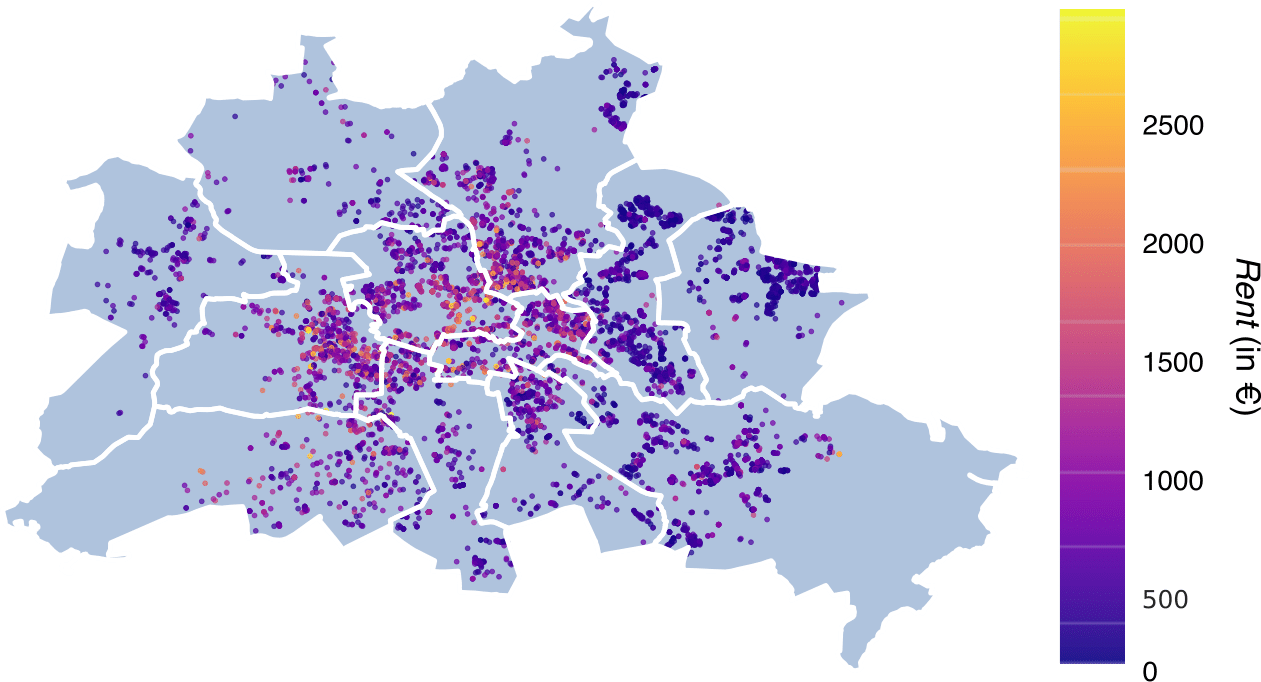}\label{fig:berlin_districts_rent}}
	\caption{Heatmap of monthly apartment costs across districts in Berlin.}
	\label{fig:berlin_districts}
\end{figure*}

\begin{figure*}
	\captionsetup{position=top}
	\begin{tabular}{ccc}
		\subfloat[$\boldsymbol{\mathit{Area}}$]{\includegraphics[width = 0.25\linewidth]{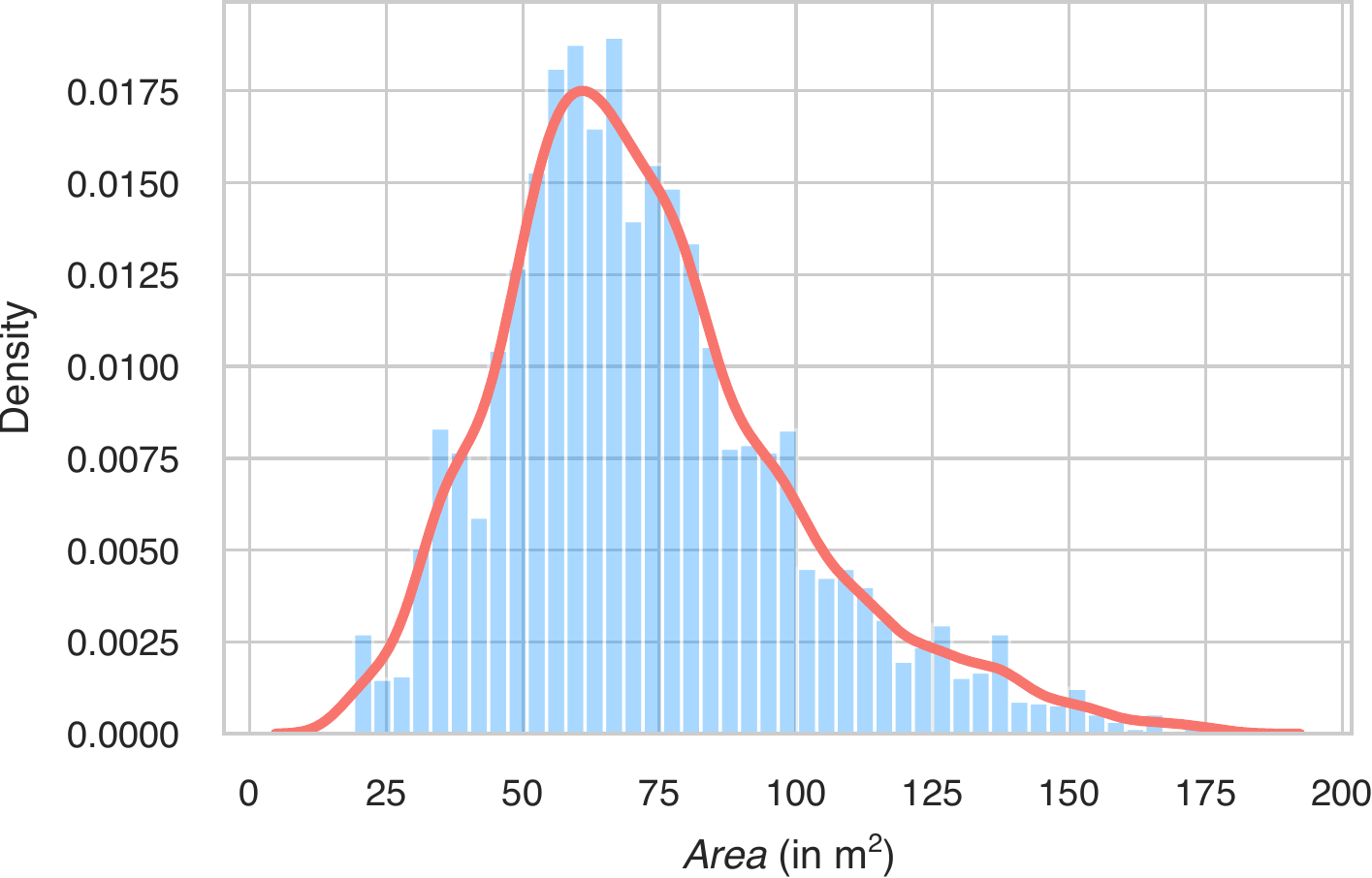}}                 &
		\subfloat[$\boldsymbol{\mathit{Rooms}}$]{\includegraphics[width = 0.25\linewidth]{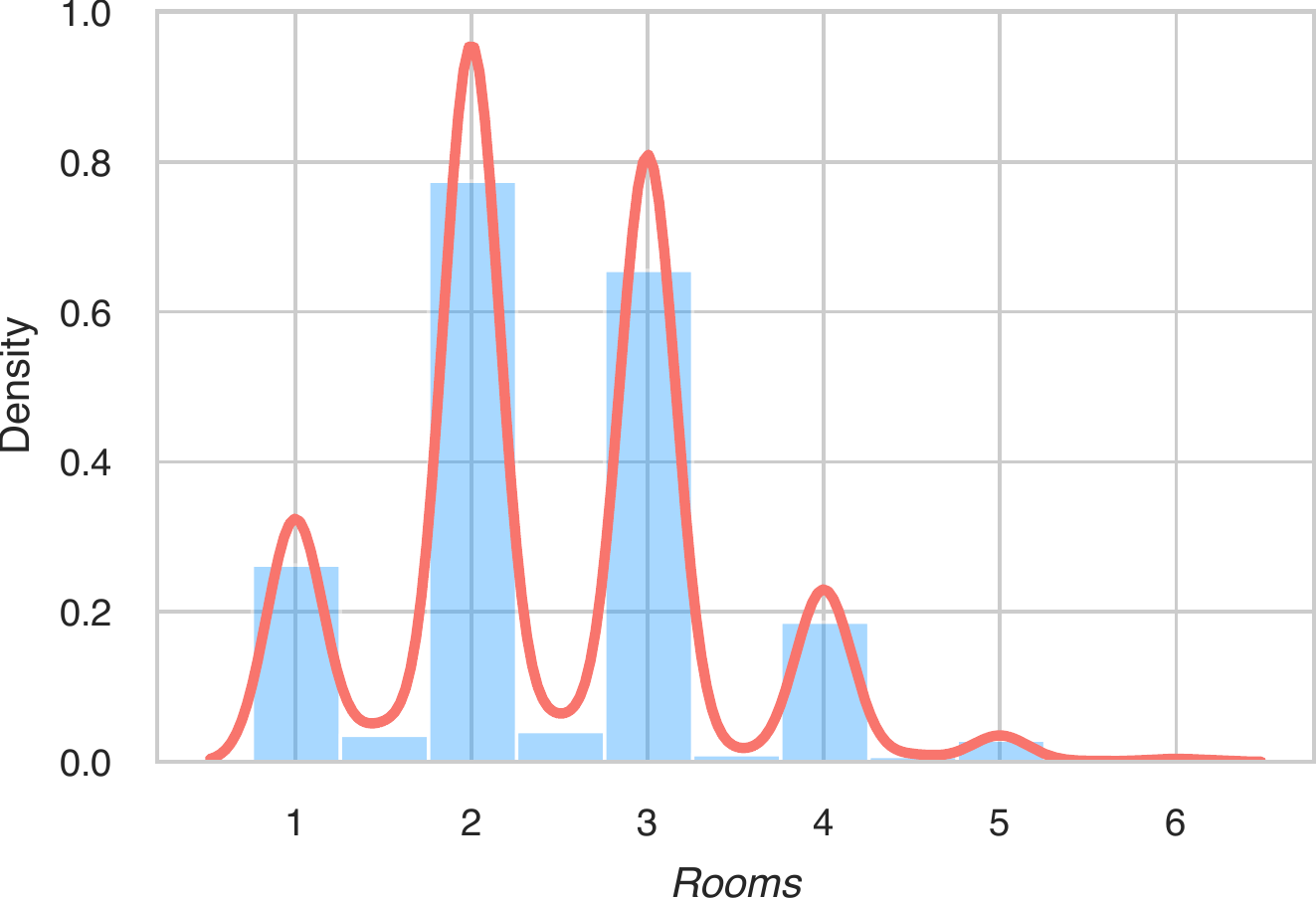}}               &
		\subfloat[$\boldsymbol{\mathit{Floor}}$]{\includegraphics[width = 0.25\linewidth]{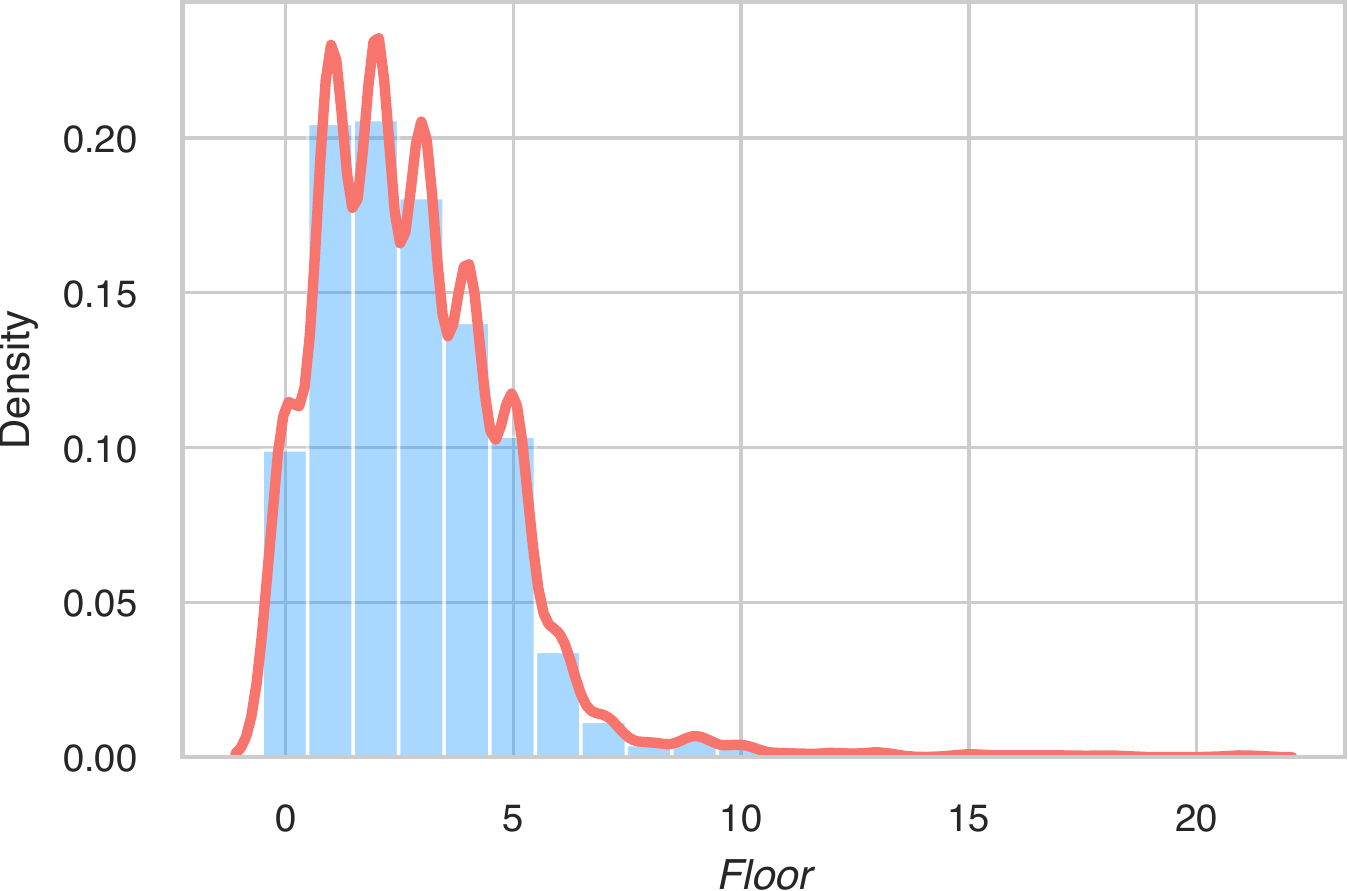}}                 \\
		\subfloat[$\boldsymbol{\mathit{TopFloor}}$]{\includegraphics[width = 0.25\linewidth]{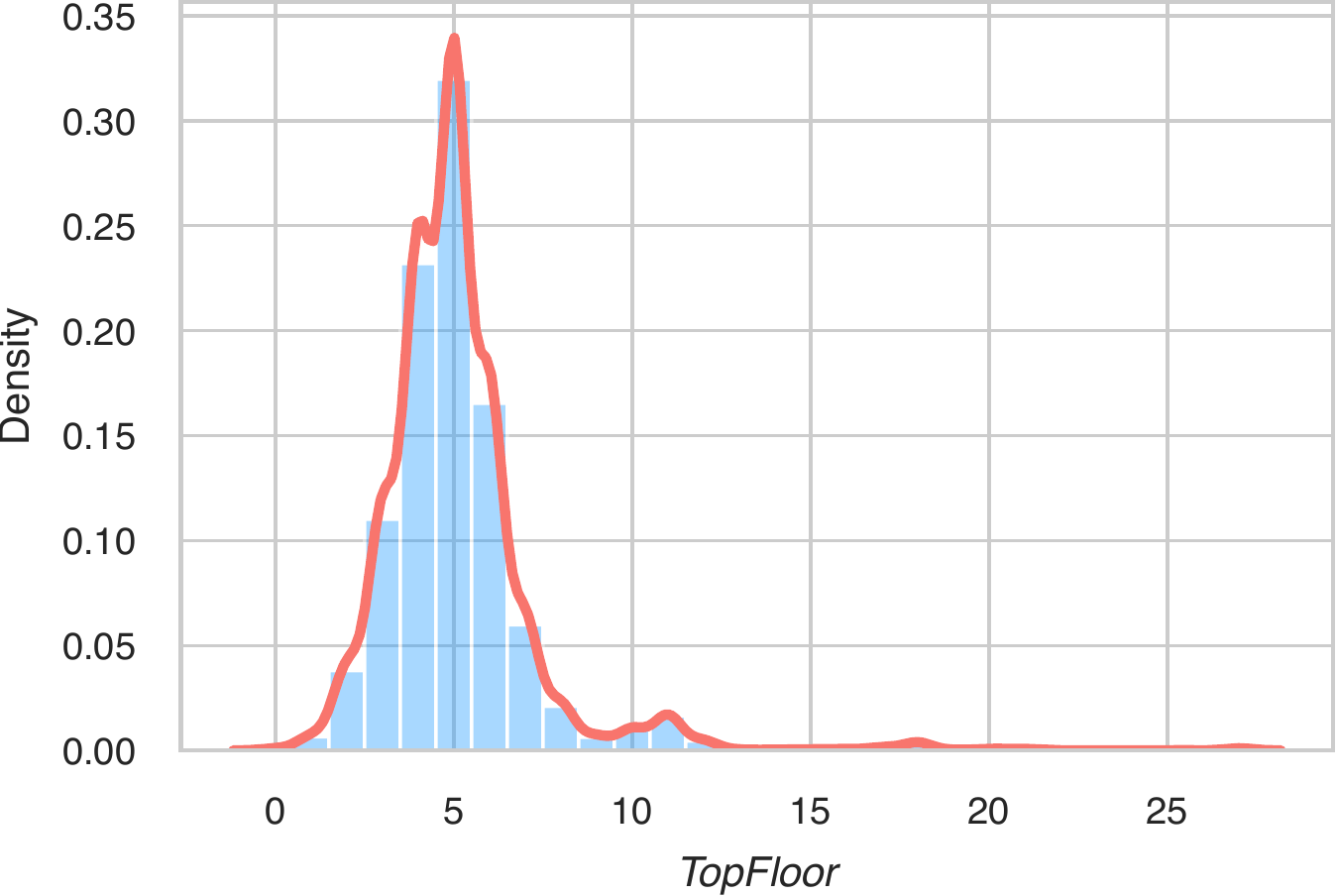}}        &
		\subfloat[$\boldsymbol{\mathit{CityDistance}}$]{\includegraphics[width = 0.25\linewidth]{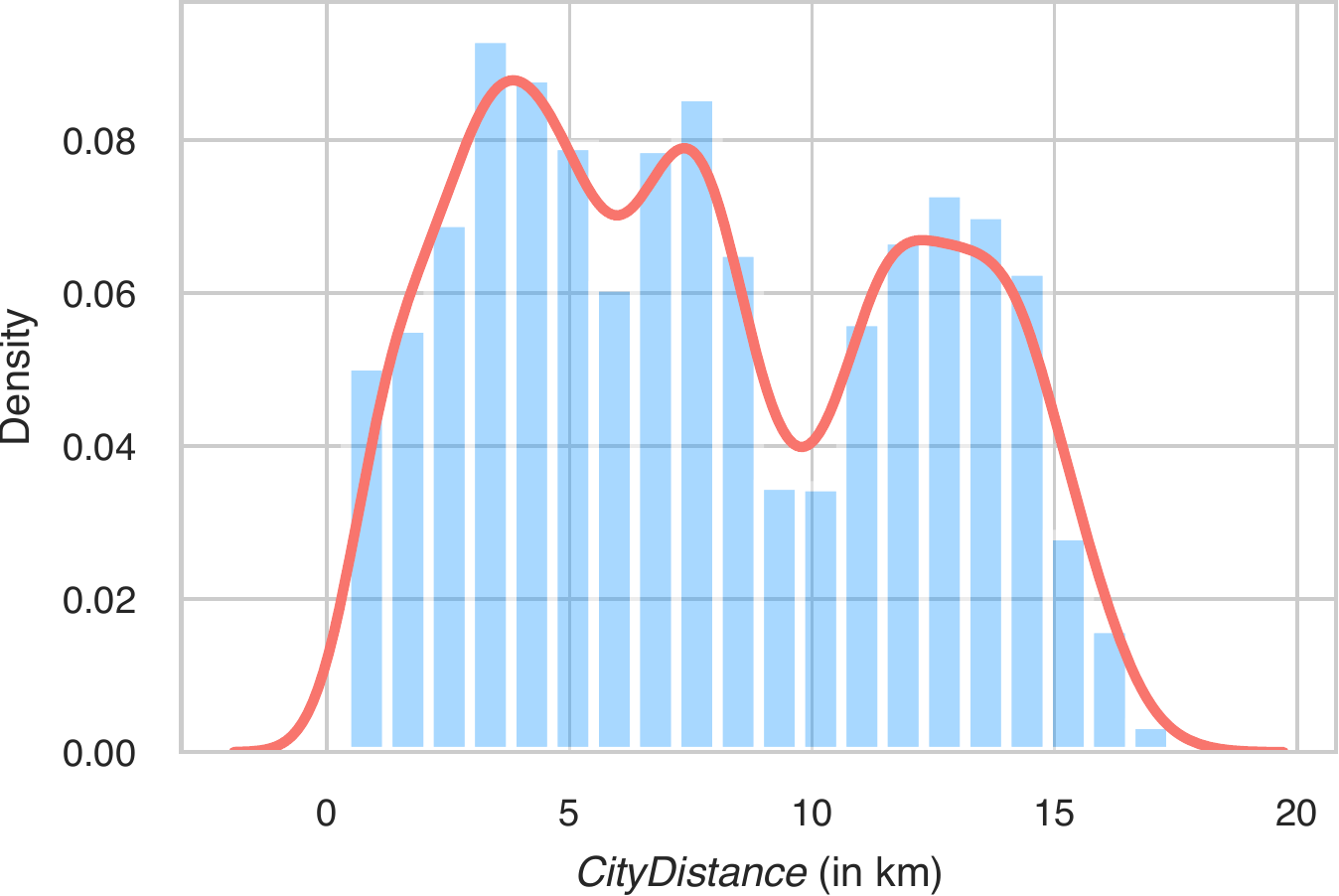}} &
		\subfloat[$\boldsymbol{\mathit{YearBuilt}}$]{\includegraphics[width = 0.25\linewidth]{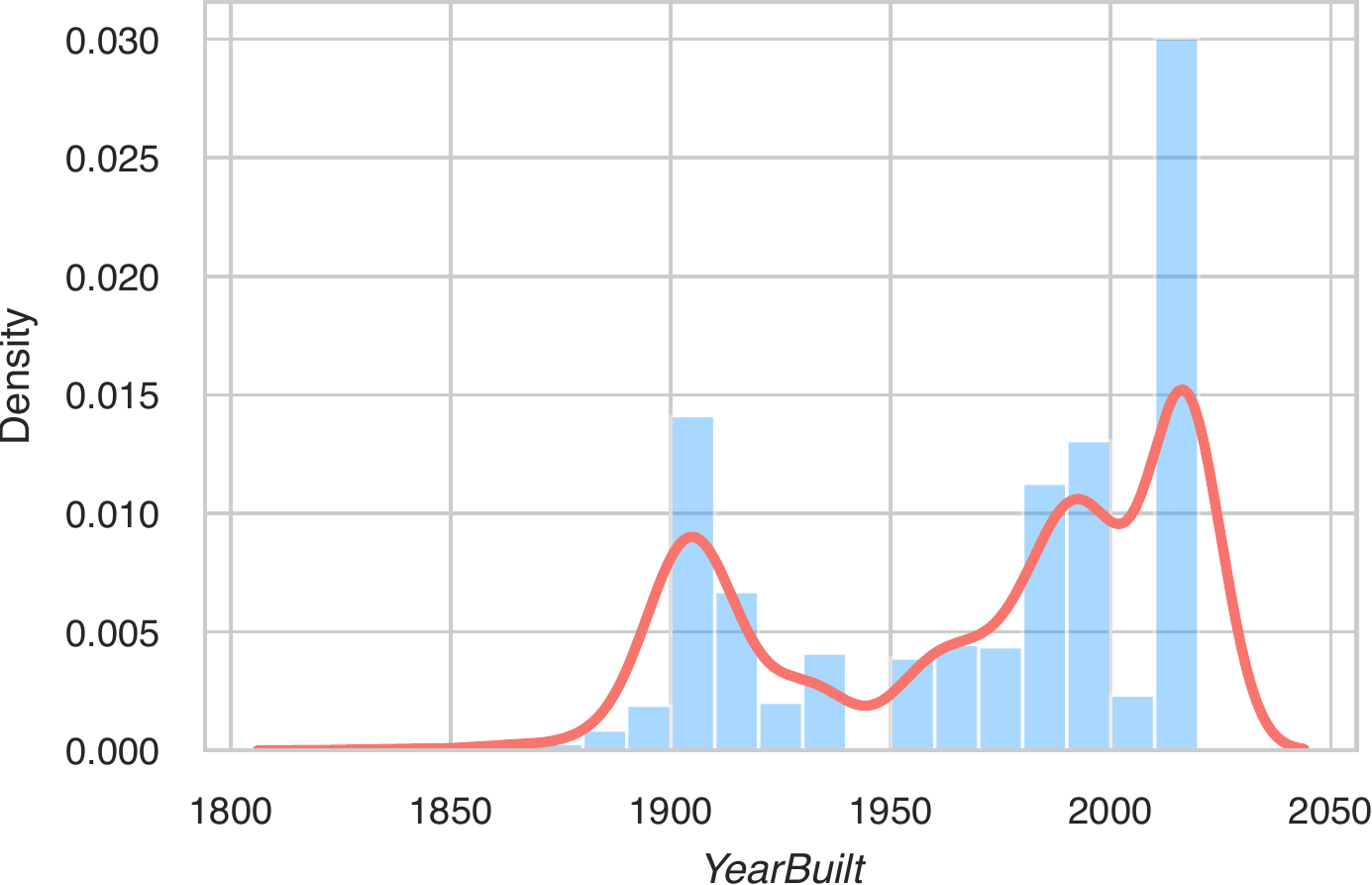}}        \\
	\end{tabular}
	\caption{Distribution of explanatory variables.}
	\label{fig:distributions}
\end{figure*}

\section{Related Work}
\label{sec:related_work}

\subsection{Hedonic Appraisal of Real Estate Prices}
\label{subsec:hedonic_appraisal}


Traditional real estate valuation research is based on financial estate theory and constructs measures of fundamental values and price indices \citep{Krainer.2004} using a wide variety of methods, including comparison \citep{Krainer.2004}, assessed value \citep{Clapp.1992}, and time series analysis \citep{Chiang.2005}.
Yet these approaches are severely limited by their scope; if a parameter has no clear market value, it is usually ignored.
For example, proximity to the transport hubs of the city may be important for the potential renter but is not taken into account by classical models.


Hedonic models deviate from standard assumptions and take into account parameters that are not easily quantifiable in terms of the market price \citep{Sirmans.2005, Pagourtzi.2003}.
In general, hedonic analysis is concerned with the marginal changes in the target variable (\eg rent price) by the variable of interest.
While still taking the basic structural parameters into account, hedonic analysis is not constrained by a comparison between similar properties and can be extended to incorporate almost any kind of additional information.
They place greater weight on the heterogeneity of the market and try to estimate the effects of individual characteristics of the property \citep{Malpezzi.2002}. Hedonic real estate models typically show relatively high explanatory power. For example, in the German market, the hedonic model developed by \citep{Kolbe.2014} shows an $R^2$ of above \num{0.7}.
While hedonic models are frequently implemented via linear regression, previous research has demonstrated that the hedonic approach can also be used in combination with neural networks and other machine learning methods \cite[\eg][]{Peterson.2009}.


Hedonic models typically account for structural and locational characteristics of a property \citep[\eg][]{Natividade.2007}.
They may vary in accordance with the available information and research questions, which is both a benefit and a possible deficiency of the method.
The authors in \cite{Sirmans.2005} provide an overview of the parameters used in the previous works and their effects.
Popular structural categories include the size and the number of floors of a property, which mostly have a positive effect on the price.
Locational characteristics tend to contain distance measures that can have an effect in either direction.
For example, school districts mostly have a negative effect \cite{Sirmans.2005}.
More unconventional specifications of hedonic models may further include the forced nature of the sale \citep{Andersen.2016}, non-euclidean distance metrics \citep{Lu.2014}, and news sentiment \citep{Sun.2014}.

\subsection{Image Analysis in Real Estate}


In the area of real estate, computer vision and visual analytics have found varying degrees of success as tools for visualization, sentiment analysis and feature extraction.
In particular, machine learning can be used to visualize real estate data for easier understanding \citep{Sun.2013, Li.2018}.
Furthermore, previous works have used machine learning to extract information and sentiment contained within images to better explain or to predict the property price \citep{H.Ahmed.2016, Glaeser.2018, You.2017, Naumzik.2020}.
For instance, photos of surrounding areas, such as satellite images, have been used to extract neighborhood data as a part of a hedonic process \citep{Bency.2017}.
Other popular data sources are interior and exterior images of the property or neighboring objects made from the ground or with the help of satellites.
Glaeser et.\ al.\ \cite{Glaeser.2018} show that exterior and neighborhood aesthetics can enhance the explanatory power of hedonic models.

\vspace{0.2cm}
\textbf{Research on real estate floor plans: }
Previous research on real estate floor plans has primarily focused on summarizing and extracting structured information from floor plans, yet with clear differences from our study.
For example, Goncu et.\ al.\ \citep{Goncu.2015} have created an application that leverages text and geometry recognition software to convert floor plan images into a vector file with contrasting colors.
The goal is to provide a simplified, more accessible version of floor plans to sight-impaired users.
Other works have focused on facilitating automatic search of homes based on floor plans.
Sharma et.\ al.\ \cite{Sharma.2019} developed a machine learning approach that maps floor plan images or user sketches of floor plans to floor plan images in a database.
An enhanced version of this approach (``FloorNet'') was developed by \cite{Kato.2020} with the goal of making more accurate real estate recommendations for users on online real estate platforms.
The authors' findings suggest that floor plans contain additional information that is not available in the structured data, such as the way rooms are connected.
Another approach is to analyze floor plans using adjacency graphs with nodes labeled as rooms and/or corridors \citep{Hanazato.2005}.
However, this method requires extensive manual labeling and can only provide partial data contained in the floor plans \citep{Kato.2020}.

\vspace{0.2cm}
\textbf{Research gap: }
None of the above references has utilized floor plans to enhance rent appraisal in hedonic pricing models.
In this paper, we close this research gap by incorporating floor plans into hedonic models for rent price appraisal on online real estate platforms.
Unlike exterior, interior, or satellite images, which present only a part of the apartment and are external to the property itself, we expect floor plans to be an integral part of property valuation and purchasing decision-making.
Floor plans contain information about relative dimensions, positioning of rooms and utilities, and thus may carry price-relevant information in the real estate market.
To the best of our knowledge, our research is the first to demonstrate that harnessing floor plans can enhance real estate appraisal in hedonic pricing models -- even after controlling for structured apartment characteristics such as size and location.

\section{Data}

\subsection{Apartment Listings}

The data for this study was extracted from ImmobilienScout24.de, the largest German online real estate aggregator.
At the end of 2019, ImmobilienScout24 contained \num{91415} active rent listings across Germany, with approximately \num{44} million visitors per month.

We implemented a Python-based web crawler to download and store information about apartments in the city of Berlin in a database format.
The crawler was active for three weeks at the end of 2019, during which it gathered a total number of \num{15604} observations. Each observation represents an apartment in Berlin from an active or archived listing on ImmobilienScout24. Our dataset includes apartment listings between mid-2017 and end of 2019 (\ie, spans two and a half years). For each apartment, the crawler collected a raw image showing the floor plan of the apartment, the monthly rent price, the monthly rent price per $\text{m}^2$, the city district in which the apartment is located, and \num{36} fields with categorical and numerical information about the apartment (\eg, apartment size, etc.)

\subsection{Variable Definitions}
\label{sub:var_def}

\textbf{Dependent variables:} The target variables in our study are two measures of the monthly costs of an apartment:
\begin{itemize}
	\item $\mathit{Rpms}$: The monthly rent price (in \euro) per ${\text{m}^2}$
	\item $\mathit{Rent}$: The total monthly rent price (in \euro) of an apartment
\end{itemize}

\vspace{0.2cm}
\textbf{Explanatory variables:} Our hedonic rent price model accounts for the following apartment characteristics from previous works:
\begin{itemize}
	\item $\mathit{Area}$: The total size of the apartment in ${\text{m}^2}$
	\item $\mathit{Rooms}$: The total number of rooms in the apartment
	\item $\mathit{Floor}$: The floor at which the apartment is located
	\item $\mathit{TopFloor}$: The number of floors of the building in which the apartment is located
	\item $\mathit{CityDistance}$: The distance (in km) from the center of the city to the apartment
	\item $\mathit{YearBuilt}$: The construction year of the building in which the apartment is located
\end{itemize}

\vspace{0.2cm}
\textbf{Control variables: }
Our dataset further includes 12 locational dummies that provide information about the city district, and a set of 30 additional control variables in the form of categorical dummies that provide fine-grained information about contractual and structural characteristics of an apartment (\eg whether the apartment offers access to a parking lot).

\subsection{Summary Statistics}

\begin{figure}
	\captionsetup{position=top}
	\begin{tabular}{cc}
		\subfloat[$\mathit{\boldsymbol{Rpms}}$]{\includegraphics[width = 0.45\linewidth]{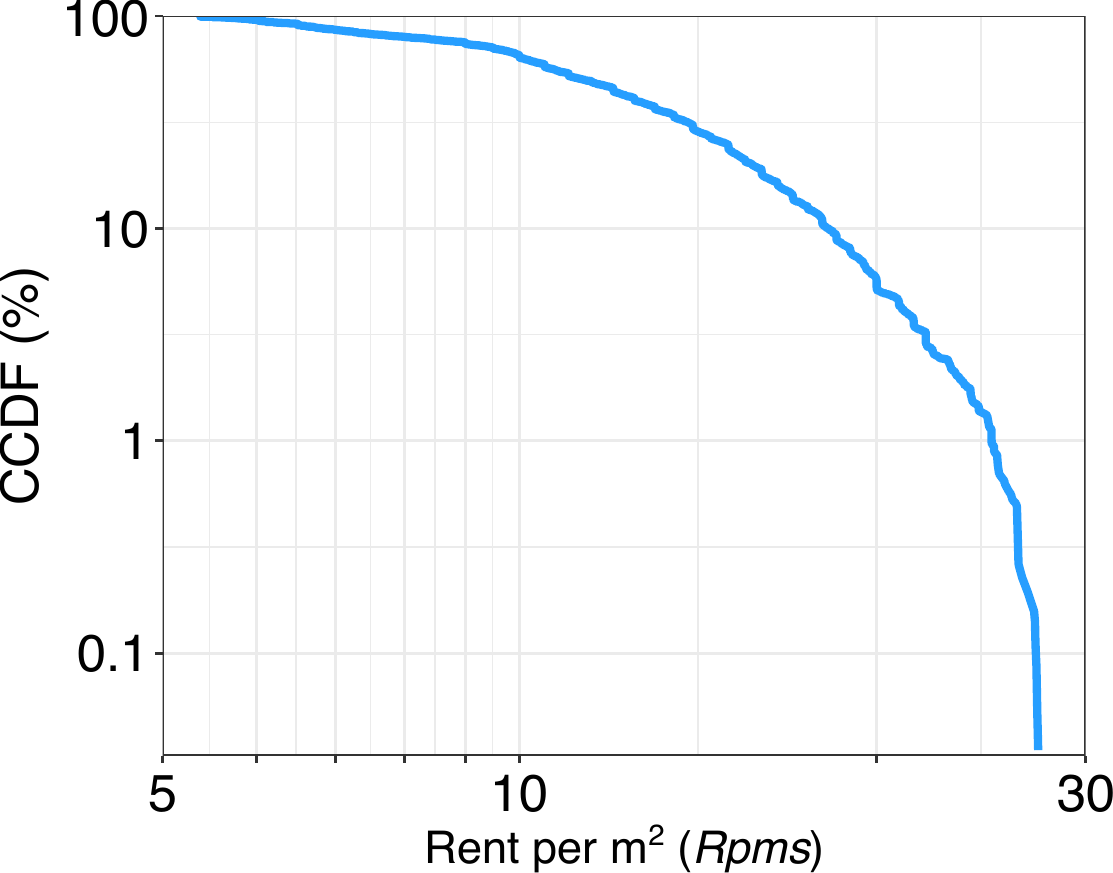}}
		\subfloat[$\mathit{\boldsymbol{Rent}}$]{\includegraphics[width = 0.45\linewidth]{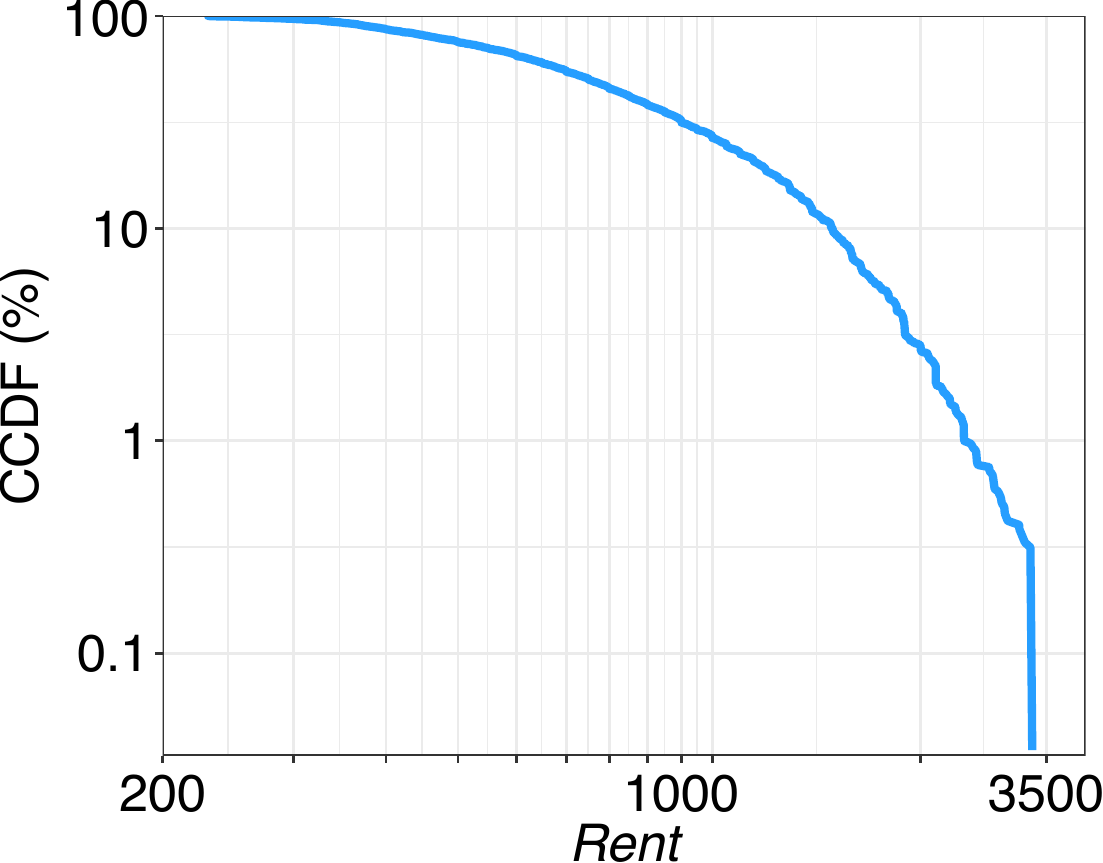}} &
	\end{tabular}
	\caption{Complementary cumulative distribution functions (CCDFs) for $\boldsymbol{\mathit{Rpms}}$ and $\boldsymbol{\mathit{Rent}}$.}
	\label{fig:ccdf}
\end{figure}

\textbf{Dependent variables: }
\Cref{tbl:descriptive_statistics} shows summary statistics, whereas Fig.~\ref{fig:ccdf} plots the complementary cumulative distributions (CCDFs) for the dependent variables.
The total rent prices ($\mathit{Rent}$) in our data range from \euro{230} per month to almost \euro{3000} with a standard deviation of \num{476.80}.
The rent per $\text{m}^2$ ($\mathit{Rpms}$) ranges from \num{5.38} to \num{27.80} with a standard deviation of \num{4.32}.
Fig.~\ref{fig:berlin_districts} provides an overview of the monthly apartment costs across different districts in Berlin.
Evidently, there is a higher concentration of apartments with high monthly costs near the city center of Berlin.

\vspace{0.2cm}
\textbf{Explanatory variables: }
Fig.~\ref{fig:distributions} plots the distributions of the explanatory variables in our dataset.
The size of the apartments ($\mathit{Area}$) ranges from \num{16.00} to \num{178.08} square meters with a standard deviation of \num{26.35}.
The number of rooms ranges from \num{1} to \num{6}, with most of the apartments having either \num{2} or \num{3} rooms.
The average distance between an apartment and the city center ($\mathit{CityDistance}$) is \num{7.99} km.
The variables $\mathit{TopFloor}$ and $\mathit{Floor}$ have the same minimum of \num{0} and a maximum of \num{27} and \num{26}, respectively.
The houses in our sample have been built between the years 1830 and 2020.
Most buildings are relatively new with a median construction year of 1985.

\begin{table}
	\sisetup{round-mode=places,round-precision=2,table-format=2.2}
	\vspace{-.5cm}
	\small
	\setlength{\tabcolsep}{3pt}
	\centering
	\begin{tabular}{lS[table-format=3.2]S[table-format=3.2]S[table-format=3.2]S[table-format=4.2]S[table-format=3.2]}
		\toprule
		\textbf{Variable}                       & \multicolumn{1}{c}{\textbf{Mean}} & \multicolumn{1}{c}{\textbf{Median}} & \multicolumn{1}{c}{\textbf{Min}} & \multicolumn{1}{c}{\textbf{Max}} & \multicolumn{1}{c}{\textbf{Std.~dev}} \\
		\midrule
		\multicolumn{2}{l}{\underline{Dependent variables}}                                                                                                                                                                             \\
		\quad $\mathit{Rpms}$ (in \EUR)         & 11.73                             & 11.0                                & 5.38                             & 27.80                            & 4.32                                  \\
		\quad $\mathit{Rent}$ (in \EUR)         & 847.12                            & 720.15                              & 230.19                           & 2988.0                           & 476.84                                \\
		\addlinespace
		\multicolumn{2}{l}{\underline{Explanatory variables}}                                                                                                                                                                           \\
		\quad $\mathit{Area}$ (in $\text{m}^2$) & 71.35                             & 67.41                               & 16.0                             & 178.08                           & 26.35                                 \\
		\quad $\mathit{Rooms}$                  & 2.44                              & 2.0                                 & 1.0                              & 6.0                              & 0.89                                  \\
		\quad $\mathit{Floor}$                  & 2.89                              & 3.0                                 & 0.0                              & 26.0                             & 2.29                                  \\
		\quad $\mathit{Top Floor}$              & 5.09                              & 5.0                                 & 0.0                              & 27.0                             & 2.21                                  \\
		\quad $\mathit{City Distance}$ (in km)  & 7.99                              & 7.51                                & 0.37                             & 17.54                            & 4.26                                  \\
		\quad $\mathit{Year Built}$             & \multicolumn{1}{c}{1970}          & \multicolumn{1}{c}{1985}            & \multicolumn{1}{c}{1830}         & \multicolumn{1}{c}{2020}         & 43.76                                 \\
		\bottomrule
	\end{tabular}
	\caption{Summary statistics of key variables.}
	\label{tbl:descriptive_statistics}
	\vspace{-.5cm}
\end{table}

\subsection{Cross-correlations}

\begin{figure}
	\centering
	\includegraphics[width=.8\columnwidth]{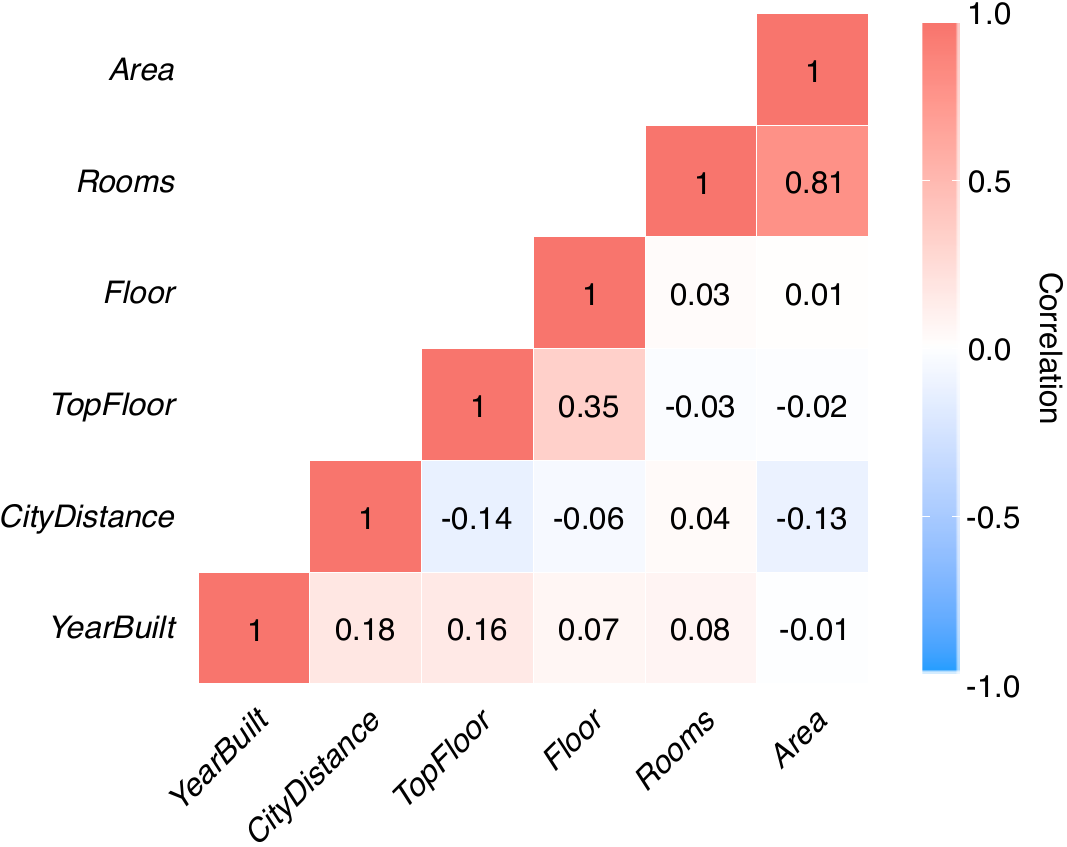}
	\caption{Cross-correlations.}
	\label{fig:correlations}
\end{figure}

Fig.~\ref{fig:correlations} provides an overview of cross-correlations between the independent variables.
Unsurprisingly, we see a positive correlation between the size of an apartment and the number of rooms (correlation of 0.81).
Likewise, we find a positive correlation of \num{0.35} between $\mathit{Floor}$ and $\mathit{TopFloor}$.
We observe a negative correlation of \num{-0.14} between $\mathit{CityDistance}$ and $\mathit{TopFloor}$, indicating that in Berlin, there is a higher concentration of tall buildings in areas close to the city center.
The correlation between $\mathit{YearBuilt}$ and $\mathit{TopFloor}$ (\num{0.16}) suggests that newer buildings tend to have more floors, while the correlation between the construction year and the distance from the city center (\num{0.18}) highlights that there has been more construction activity near the edges of Berlin in recent years.
All remaining correlations are fairly small.
Importantly, the variance inflation factors of all independent variables in our later analysis are below the critical threshold of 4.
Hence there is no evidence that multicollinearity impedes the validity of our findings.

\subsection{Extraction of Floor Plans from Images}

We perform three main preprocessing steps to transform the raw floor plans into a format that allows for further calculations.
First, we manually remove erroneous entries containing images that are not floor plans.
These include interior and exterior photos, promotional materials, and presentation slides.
Second, we exclude technically correct but unrelated floor plans.
For example, if there is a floor plan for the parking space, it is removed, so the neural network does not learn unrelated information. These filtering steps reduce our dataset to \num{9174} observations.

\begin{figure}
	\centering
	\includegraphics[width=.9\columnwidth]{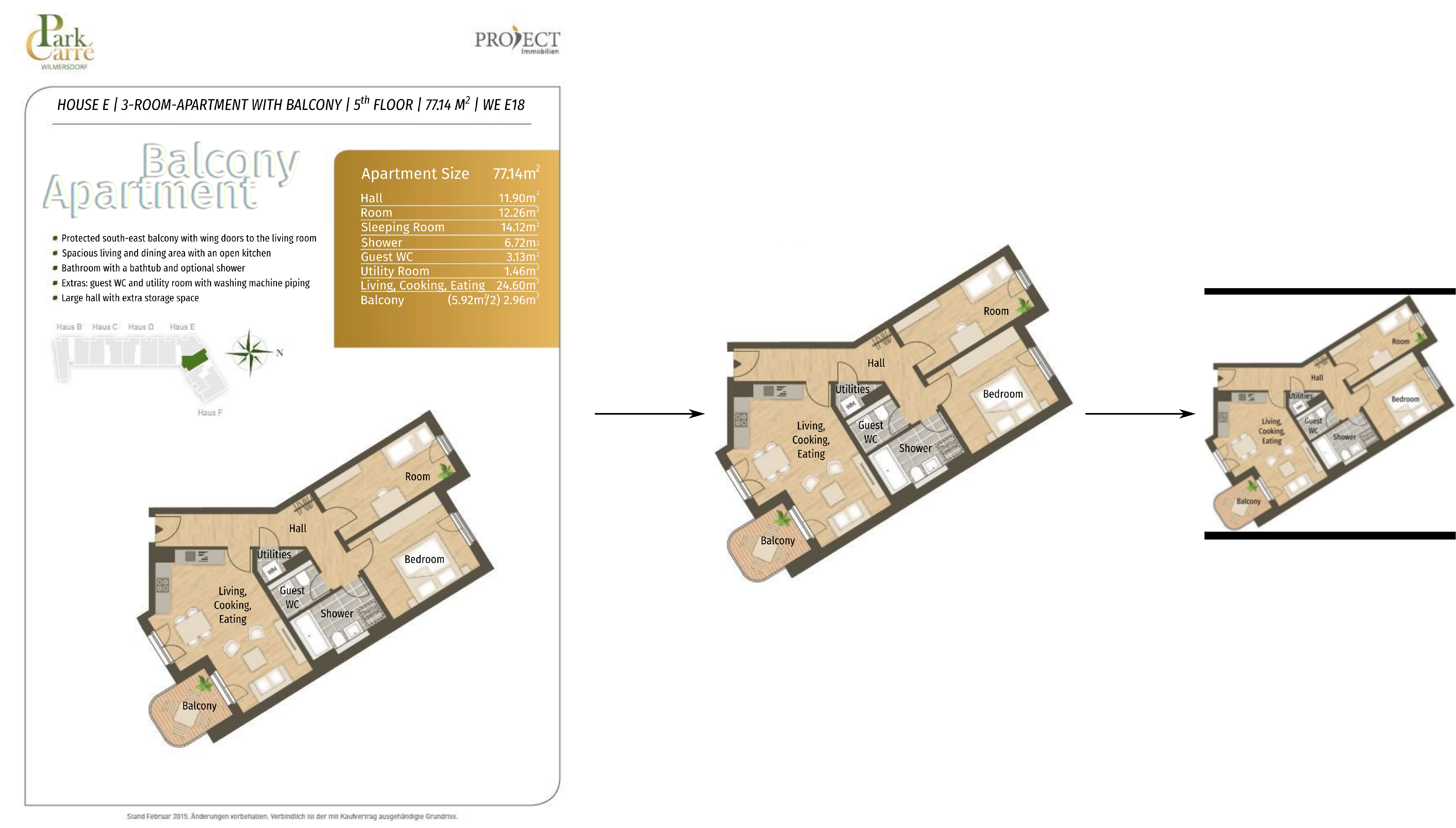}
	\caption{Example of cropping and scaling of floor plans.}
	\label{fig:scale}
\end{figure}

Once the data set is reduced to contain only the relevant entries, we modify them for the use in neural networks.
We crop every picture until only the plan remains.
Subsequently, we scale them with proportions preserved, filling empty space with pure black to prevent unnecessary calculations.
All floor plans are then adjusted to follow the Xception guidelines \citep{Chollet.2017} of min-max normalized color images with \num{299} by \num{299} pixels in size.
For this purpose, we use an ImageMagick script that scales images without shifts in proportions to avoid potential loss of information and to lighten the computational burden.
The prepared images are then vectorized, min-max normalized, and concatenated to the data frame.
The process is illustrated in Fig.~\ref{fig:scale}.

\section{Methodology}

We propose a two-stage approach to determine the hedonic price of apartment floor plans.
As a first step, we compute adjusted rent prices by regressing the monthly apartment costs on the structural and locational variables (Stage 1).
This allows us to extract the variance in rent that is unexplained by these variables and prevents our model from learning information that is already available from known features.
Second, we use convolutional neural networks (CNN) and the adjusted rent prices to predict the sentiment of the floor plans, \ie the apartment rent price after controlling for locational and structural characteristics of an apartment (Stage 2).

\subsection{Computing Adjusted Rent Prices (Stage 1)}

Stage (1) performs a hedonic regression to calculate the variance in the apartment costs that is unexplained by the structural and locational attributes of an apartment.
This approach follows previous works \cite{Dehaan.2013,Ball.2012,Farrell.2014,Naumzik.2020}, where regression residuals are used as a proxy for the proportion of variance of the dependent variable that is unexplained by known determinants.
Let $y_{i}$ denote the logarithmized rent/rent per $\text{m}^2$ belonging to listing $i = 1, \dots, n$.
We then estimate a linear model via ordinary least squares (OLS) regressing $y_i$ on the structural and locational attributes $c_{i1}, \dots, c_{iJ}$.
Formally, the model is described by
	{%
		\begin{equation}
			y_i = \beta_0 + \sum_{j=1}^J \beta_j c_{ij} + \varepsilon_i \text{,}
		\end{equation}
	}%
with intercept $\beta_0$ and error term $\varepsilon$.
The model residuals \(\tilde{y_{i}} = y_{i} - \hat{y_{i}}\) then represent adjusted rent prices that are used in a neural network model in Stage (2).
Importantly, the calculation of the adjusted prices ensures that the neural network does not merely learn to predict easily observable structural apartment characteristics.

\subsection{Floor Plan Sentiment (Stage 2)}

Stage (2) uses convolutional neural networks (CNN) to predict the sentiment of the floor plans. The input to the model is a vector representation of the preprocessed floor plan images ($299 \times 299$ pixels). The variable to predict is the adjusted rent price from Stage (1), \ie, $\tilde{y_{i}}$.
We implement our learning task using Xception \citep{Chollet.2017} developed by Google, which is the state-of-the-art in image classification \citep{Li.2018}.
Specifically, we employ a pretrained Xception model trained on a large dataset of images from ImageNet \citep{Chollet.2017} and employ a transfer learning strategy (\ie, learn a new task through the transfer of knowledge from a related task that has already been learned) to transfer the pre-trained Xception model to our floor plan prediction task.

\begin{figure}
	\centering
	\includegraphics[width=\columnwidth]{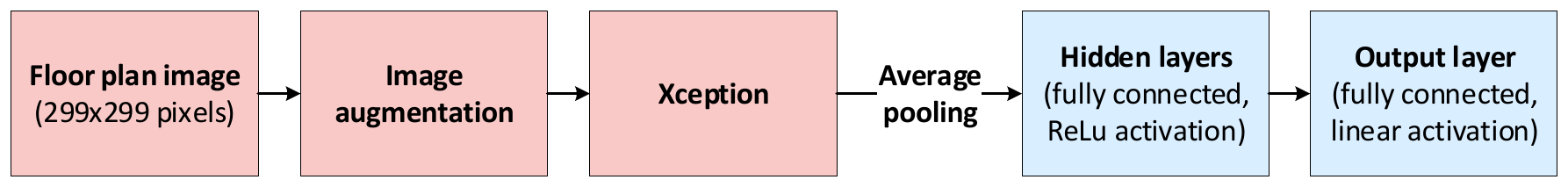}
	\caption{Illustration of model architecture.}
    \label{fig:architecture}
\end{figure}

\textbf{Architecture customization:} The pretrained Xception model uses a softmax layer as the final layer to make categorical predictions. As our goal is to use floor plan images to predict a continuous variable ($\tilde{y_{i}}$), we modify the architecture by removing the softmax layer from the pre-trained model. Further, we apply the global average pooling operation on the output layer. We then append a set of fully connected layers with ReLu activation function, followed by an output layer consisting of a single neuron (with linear activation function). The latter ensures that our model produces continuous predictions of $\tilde{y_{i}}$. The number of hidden layers and nodes are treated as hyperparameters and the optimal numbers are selected by applying a grid search using 5-fold cross-validation. Our model architecture is illustrated in Fig.~\ref{fig:architecture}.

\textbf{Image augmentation: }
To increase the number of available floor plans, we implement online image augmentation.
In particular, each floor plan has a probability of being mirrored or rotated by an arbitrary number of radians.
While this does not change any information in the image itself, it expands the number of images available to the neural network, which may help it to better understand different features and combat overfitting.

\textbf{Training: }
We implement the CNN in Keras with GPU-accelerated TensorFlow as backend. Formally, we train\footnote{We use relu activations for the hidden layers, linear activation for the output layer, and Nadam optimizer. We use a validation split of \SI{20}{\percent} to determine the optimal stopping point. We optimize the number of layers, the number of neurons per layer for each layer, learning rate, and betas via grid search.} the function \(f_{\theta}\) by mapping the image vector representations of the (augmented) floor plans \(x_{i}\) to the training labels \(\tilde{y_{i}}\), resulting in an estimated parametrization \(\hat{\theta}\).
The CNN is trained by minimizing the loss between the trained function $f_{\theta}$ and the price residual $\tilde{y_{i}}$.
In the following, we refer to the resulting predictions as the sentiment of the floor plans ($\mathit{FloorPlanSentiment}$).
These predictions are later used in (i)~a hedonic regression model to evaluate the {explanatory power} of floor plans, and (ii)~as an additional feature for {out-of-sample prediction} of rent prices.

\section{Empirical Analysis}

We now empirically investigate to what extent an automated visual analysis of floor plans can enhance real estate appraisal on online real estate marketplaces.
First, we link to earlier research by studying the hedonic pricing value of floor plans in a hedonic regression model.
Second, we evaluate the gains in prediction performance when incorporating floor plans.

\subsection{Hedonic Regression Analysis}
\label{sub:regression}

\textbf{Model specification: }
We apply a log-linear regression model to analyze the role of floor plans for hedonic rent price appraisal.
Regression models are generally regarded as an explanatory approach with the ability to document statistical relationships and, in particular, estimating effect sizes \citep{Breiman.2001}.
Furthermore, log-linear regression models are widely used to estimate the hedonic rent price models in the real estate sector (\cf Sec.~\ref{sec:related_work}).
This modeling approach allows us to interpret the model coefficients and statistically test the association between floor plans and rent prices.

The key explanatory variable in our hedonic regression model is the sentiment of the floor plan images ($\mathit{FloorPlanSentiment}$), which we calculate based on the methodology described in the previous section.
Importantly, we use 5-fold cross-validation to predict the $\mathit{FloorPlanSentiment}$.
Hence, \emph{FloorPlanSentiment} refers to out-of-sample predictions. 
The complete hedonic regression model is:
\begin{equation}
	\begin{aligned}
		y_{i} = & \, \beta_0 + \beta_{1}Floor Plan Sentiment_{i} + \sum_{j=1}^{J}{\gamma_{j}c_{ij}} + \varepsilon_i \text{,}
	\end{aligned}
	\label{eq:full}
\end{equation}
\noindent where the dependent variable $y_{i}$ is the log of either $\mathit{Rpms}$ or $\mathit{Rent}$, $\beta_{0}$ is the intercept, and the coefficients $\gamma$ gauge the effects of the explanatory and control variables.
The coefficient $\beta_1$ captures the marginal effect of the $\mathit{FloorPlanSentiment}$.
This is our parameter of interest as it measures the contribution of the visual design of floor plans on the monthly costs of an apartment.
Note that, by combining floor plans and structural/locational in a joint regression model, we can isolate the \textbf{marginal} effect of floor plans on the monthly costs of an apartment.

\vspace{0.2cm}
\textbf{Estimation: }
We follow previous works and estimate Eq.~\ref{eq:full} via ordinary least squares (OLS).
Since rent prices tend to be log-normally distributed, we log-transform the dependent variables.
For the sake of interpretability, we also $z$-standardize all variables, so that the regression coefficients measure the relationship with the dependent variable measured in standard deviations.

\begin{figure}
	\centering
	{\includegraphics[width=\columnwidth]{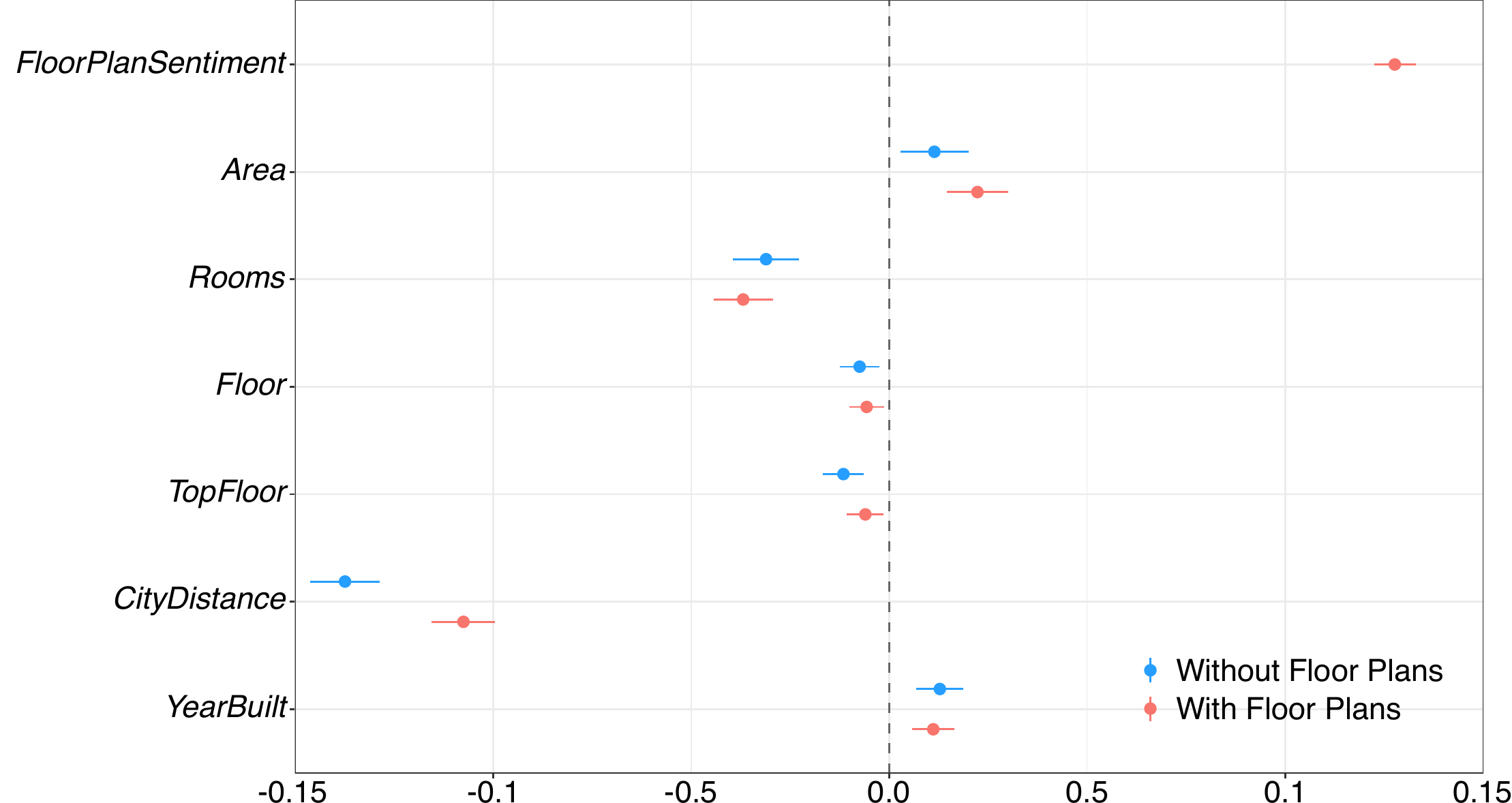}\label{fig:coeff_plot}}
	\caption{Standardized parameter estimates and \SI{95}{\percent} confidence intervals.}
	\label{fig:regression_coefficients}
\end{figure}

\vspace{0.2cm}
\textbf{Coefficient estimates for rent per $\boldsymbol{\text{m}^2}$ ($\boldsymbol{\mathit{Rpms}}$): }
The parameter estimates in Fig.~\ref{fig:regression_coefficients} show that the visual design of floor plans has significant explanatory power regarding rent prices.
The coefficient for $\mathit{FloorPlanSentiment}$ has the largest positive effect size, and is statistically significant (coef: \num{0.128}; $p<0.001$).
A one standard deviation change in $\mathit{FloorPlanSentiment}$ is estimated to increase the rent price by \SI{13.65}{\percent}.
The largest negative effect on $\mathit{Rpms}$ is estimated for $\mathit{CityDistance}$ (coef: \num{-0.108}; $p<0.001$).
An increase in standard deviation in $\mathit{CityDistance}$ reduces the price per $\text{m}^2$ by \SI{10.23}{\percent}.
We further find statistically significant positive effects for $\mathit{YearBuilt}$ and $\mathit{Area}$.
In contrast, higher values for $\mathit{Rooms}$, $\mathit{Floor}$, and $\mathit{TopFloor}$, have a negative effect on the rent price per $\text{m}^2$.

For comparison, Fig.~\ref{fig:regression_coefficients} also reports the results for a baseline model without $\mathit{FloorPlanSentiment}$.
All coefficients remain relatively stable which confirms the validity of our results.
We again find that apartments that are smaller, located at higher floors, and farther away from the city center are associated with a lower rent per $\text{m}^2$.
Apartments with fewer rooms are associated with higher rent per $\text{m}^2$.

\vspace{0.2cm}
\textbf{Control variables: }
Our regression model controls for 30 additional structural and locational apartment characteristics (see Fig.~\ref{fig:berlin_districts}).
The corresponding estimates are omitted from Fig.~\ref{fig:regression_coefficients} for the sake of brevity.
In short, more preferable contractual clauses (e.g.\ non-coal heating system, the allowance of pets or recent renovation) result in higher rent prices.
We further observe location effects, \ie city district dummies that are statistically significant.

\vspace{0.2cm}
\textbf{Goodness-of-fit: }
We calculated the adjusted-$R^2$ for each model, resulting in relatively high values of \num{0.622} for the baseline hedonic model without floor plans.
If we additionally include floor plans, the adjusted-$R^2$ increases to \num{0.697}.
Evidently, our model specification accounts for a large proportion of the variations in the dependent variable and $\mathit{FloorPlanSentiment}$ has significant explanatory power.
This observation is supported by the difference in the AIC scores for the model with and without $\mathit{FloorPlanSentiment}$.
For both dependent variables, the difference is greater than 10, indicating strong support for the corresponding candidate models \citep{Burnham.2004}.
Therefore, the model that incorporates the floor plans is to be preferred.

\vspace{0.2cm}
\textbf{Analysis of total rent ($\mathit{\boldsymbol{Rent}}$): } The estimates for the regression with the total rent ($Rent$) as dependent variable are omitted for brevity.
In short, the coefficients are consistent with those for the rent price per $\text{m}^2$ ($\mathit{Rpms}$).
Also here, the coefficient for the $\mathit{FloorPlanSentiment}$ variable is statistically significant at the \SI{1}{\percent} statistical significance level.
The regression results again suggest that the sentiment of the floor plans explains a considerable amount of the variations in rent prices after controlling for structural and locational characteristics of an apartment.
The coefficient for $\mathit{FloorPlanSentiment}$ is again among the variables with the largest effect sizes.
In other words, floor plans contain price relevant information in online real estate listings.

\subsection{Prediction Performance}

Next, we examine whether floor plans are useful regarding out-of-sample prediction of rent prices.
As mentioned in Sec.~\ref{subsec:hedonic_appraisal}, hedonic analysis is not limited to linear regression and can be used in combination with machine learning methods.
In the following, we compare the predictive value of incorporating floor plans as an additional predictor for three models: log-linear OLS regression (acting as a benchmark), boosted decision tree ensembles (CatBoost), and deep neural networks.
To evaluate the predictive value of floor plans, we split our dataset into training and test sets and compare the effect of incorporating floor plans on the out-of-sample prediction performance of the models.
Specifically, we use 5-fold cross-validation to calculate mean squared error (MSE) and mean absolute error (MAE).

\vspace{0.2cm}
\textbf{Prediction of rent per ${\boldsymbol{\text{m}^2}}$: }
\Cref{tbl:results_ppms} reports the out-of-sample prediction performance for the rent per $\text{m}^2$ ($Rpms$).
Incorporating floor plans yields improvements for all considered out-of-sample performance metrics.
A log-linear hedonic regression model yields an MSE of {0.0484} and an MAE of {0.1565}.
Incorporating \emph{FloorPlanSentiment} reduces the MSE by \SI{11.71}{\percent} and the MAE by \SI{13.93}{\percent}.
Hedonic models trained with machine learning methods yield further improvements.
A hedonic model trained with neural networks yields the lowest out-of-sample prediction error (MSE of {0.0014} and MAE of {0.0214}).
Also here, incorporating floor plans yields a substantially lower error.
The MSE is reduced by \SI{10.56}{\percent}, whereas the MAE is reduced by \SI{4.21}{\percent}.
For the decision tree ensembles (CatBoost), we find a reduction of \SI{4.29}{\percent} in MSE and a reduction in MAE of \SI{1.53}{\percent}. We also conducted paired $t$-tests to assess the statistical significance of the reductions in prediction error between the machine learning models with vs.\ without floor plans. 
We find that incorporating floor plan yields statistically significantly reduced prediction errors for all considered machine learning methods, \ie, log-linear regression ($p < 0.001$), CatBoost ($p < 0.05$), and deep neural networks ($p < 0.001$).

\begin{table}
	\sisetup{round-mode=places,round-precision=5,input-symbols=()}
	\centering
	\resizebox{\columnwidth}{!}{%
		\begin{tabular}{lcS[table-format=1.4]S[table-format=1.4]}
			\toprule
			\textbf{Method}                                           & \textbf{Floor Plans} & \multicolumn{1}{c}{\textbf{MSE}} & \multicolumn{1}{c}{\textbf{MAE}} \\
			\midrule
			Benchmark: Log-Linear Regression                          & \xmark               & {0.04984}                        & {0.1565}                         \\
			\hline
			\addlinespace
			\textbf{Hedonic models}                                                                                                                                \\
			\quad Log-Linear Regression (with floor plans)            & \checkmark           & {0.04400}                        & {0.1347}                         \\
			\quad CatBoost                                            & \xmark               & {0.00140}                        & {0.0262}                         \\
			\quad CatBoost (with floor plans)                         & \checkmark           & {0.00134}                        & {0.0258}                         \\
			\quad Deep Neural Networks                                & \xmark               & {0.00142}                        & {0.0214}                         \\
			\quad \textbf{Deep {Neural Networks (with floor plans) }} & \checkmark           & \textbf{0.00127}                 & \textbf{0.0205}                  \\
			\bottomrule
		\end{tabular}
	}
	\caption{Out-of-sample prediction performance for rent per $\boldsymbol{\text{m}^2}$ ($\boldsymbol{\mathit{Rpms}}$).
		The lowest values of mean squared error (MSE) and mean absolute error (MAE) are highlighted in bold.} %
	\label{tbl:results_ppms}
	\vspace{-.5cm}
\end{table}

\vspace{0.2cm}
\textbf{Prediction of total rent: }
\Cref{tbl:results_rent} reports the performance for the prediction of the total rent price ($\mathit{Rent}$).
All considered models show a significant reduction of the prediction error when including floor plans.
The MSE of log-linear regression, decision tree ensemble, and deep neural networks reduce by \SI{6.05}{\percent}, \SI{4.27}{\percent}, and \SI{5.08}{\percent}.
We also observe a lower MAE for log-linear regression ($-$\SI{3.92}{\percent}), decision tree ensemble ($-$\SI{2.03}{\percent}), and deep neural networks ($-$\SI{1.97}{\percent}).
Analogous to the analysis of the rent per $\text{m}^2$, paired $t$-tests show that the reductions in prediction error between the models with vs.\ without floor plans are statistically significant for each considered machine learning method. Altogether, these results demonstrate that floor plans not only feature significant explanatory power but also serve as highly relevant features for predictive purposes.

\begin{table}
	\sisetup{round-mode=places,round-precision=5,input-symbols=()}
	\centering
	\resizebox{\columnwidth}{!}{%
		\begin{tabular}{lcSS}
			\toprule
			\textbf{Method}                                           & \textbf{Floor Plans} & \multicolumn{1}{c}{\textbf{MSE}} & \multicolumn{1}{c}{\textbf{MAE}} \\
			\midrule
			Benchmark: Log-Linear Regression                          & \xmark               & {0.05078}                        & {0.1529}                         \\
			\hline
			\addlinespace
			\textbf{Hedonic models}                                                                                                                                \\
			\quad Log-Linear Regression (with floor plans)            & \checkmark           & {0.04771}                        & {0.1459}                         \\
			\quad CatBoost                                            & \xmark               & {0.00117}                        & {0.0246}                         \\
			\quad CatBoost (with floor plans)                         & \checkmark           & {0.00112}                        & {0.0241}                         \\
			\quad Deep Neural Networks                                & \xmark               & {0.00118}                        & {0.0203}                         \\
			\quad \textbf{Deep {Neural Networks (with floor plans) }} & \checkmark           & \textbf{0.00112}                 & \textbf{0.0199}                  \\
			\bottomrule
		\end{tabular}
	}
	\caption{Out-of-sample prediction performance for rent ($\boldsymbol{\mathit{Rent}}$). The lowest values of mean squared error (MSE) and mean absolute error (MAE) are highlighted in bold.} %
	\label{tbl:results_rent}
	\vspace{-.5cm}
\end{table}

\subsection{Sensitivity Analysis \& Robustness Checks}

\textbf{Analysis on data subsets: } We repeated our analysis on multiple subsets of the data.
While the increase in explanatory power stemming from the floor plans remained significant for every tested subset, we observed a statistically significant more pronounced effect for smaller apartments, as well as for older houses.

\vspace{0.2cm}
\textbf{Model specification: }
We repeated our analysis using a number of alternative model specifications, including, but not limited to CNN models (Xception, VGG16, ResNet101V2, DenseNet, EfficientNet), alternative input image formats (color, gray scale, preserved ratios, fill methods), alternative parameter configurations of the neural network (model concatenation, input concatenation, output concatenation, depths), and alternative variable specifications (normalization methods, log-specifications).
Additionally, we tested the inclusion of the images directly alongside the structured data and as a standalone feature.
In total, we have tested more than \num{50} various permutations yielding robust findings, \ie significant improvements in explanatory power and prediction performance when incorporating floor plans into hedonic price models for rent price appraisal.

\vspace{0.2cm}
\textbf{Additional robustness checks: }
We conducted additional checks to validate the robustness of our hedonic regression model: (1)~We calculated variance inflation factors for all independent variables in our hedonic regression models and found that all remain below the critical threshold of four.
(2)~We tested alternative model specifications in which naturally correlated variables such as size and number of rooms of an apartment are iteratively added one by one and tested for statistical significance after each iteration (\ie, stepwise regression).
(3)~We controlled for outliers in the dependent variables.
(4)~We added quadratic terms for each explanatory variable to our hedonic regression models.
In all cases, our results are robust and consistently support our findings.

\subsection{Exemplary Apartment Listings}

\begin{figure*}%
	\captionsetup{position=top}
	\centering
	\vspace{-.1cm}
	\subfloat[Positive price adjustment]{\includegraphics[width=.21\linewidth]{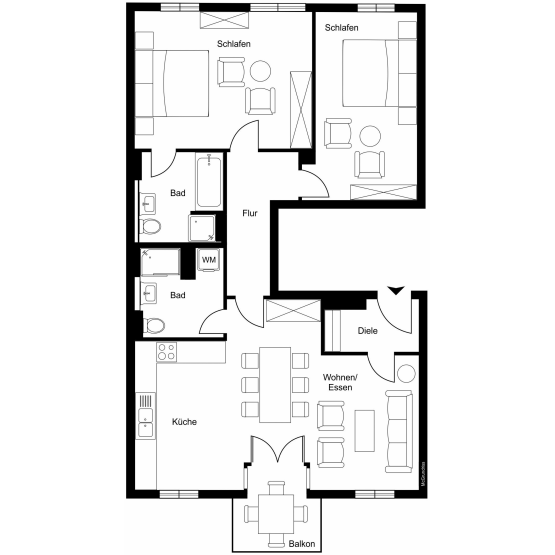}\label{fig:most_diff_a}}
	\hspace{0.35cm}
	\subfloat[Positive price adjustment]{\includegraphics[width=.21\linewidth]{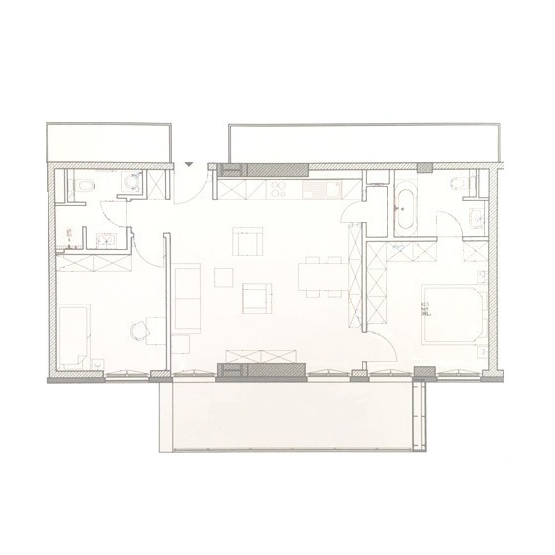}\label{fig:most_diff_b}}
	\hspace{0.35cm}
	\subfloat[Negative price adjustment]{\includegraphics[width=.21\linewidth]{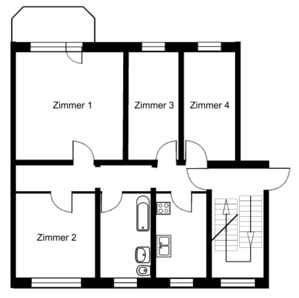}\label{fig:most_diff_c}}
	\hspace{0.35cm}
	\subfloat[Negative price adjustment]{\includegraphics[width=.21\linewidth]{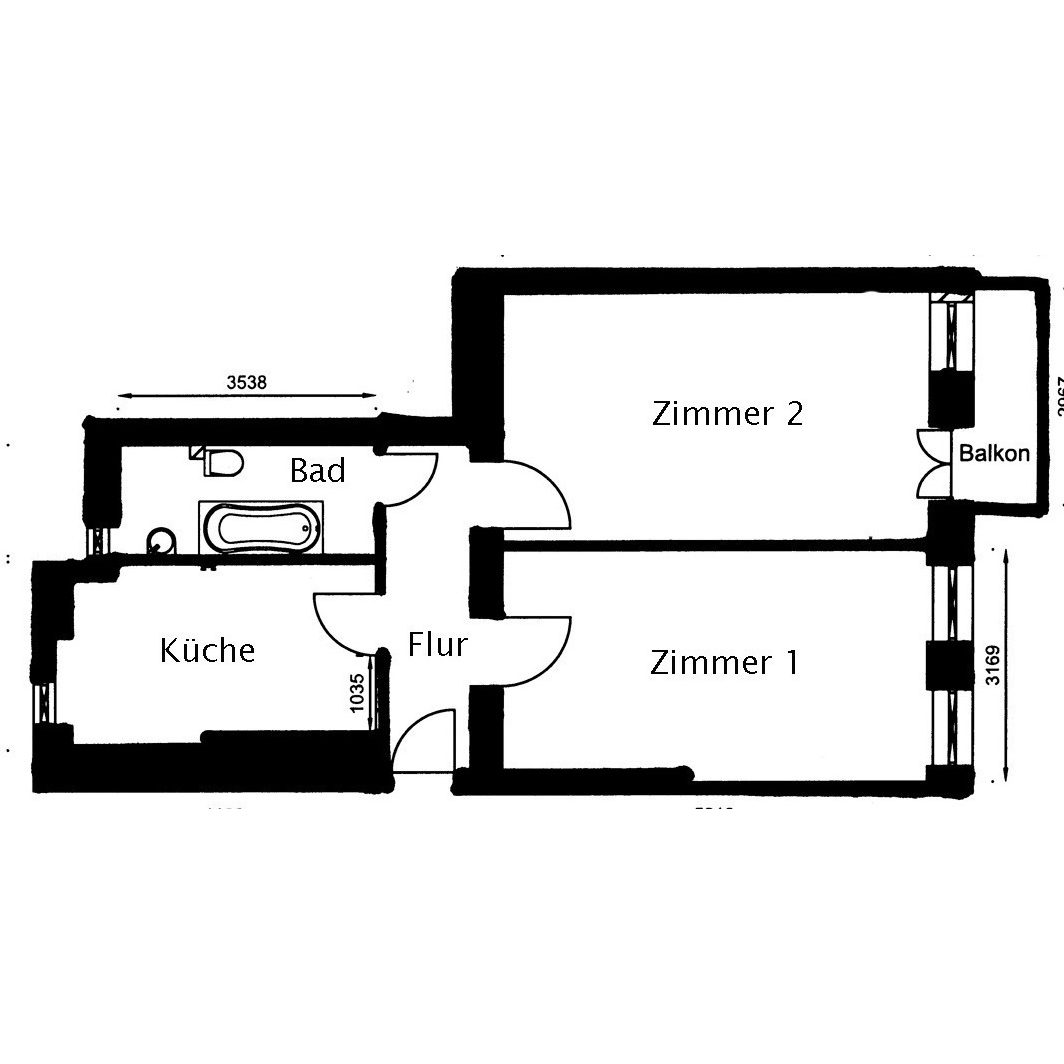}\label{fig:most_diff_d}}
	\caption{Examples of floor plans with positive and negative effects on rent prices.}
	\label{fig:diff}
\end{figure*}

We now explore apartment listings for which floor plans are particularly informative for rent price appraisal on online real estate platforms.
Fig.~\ref{fig:diff} shows four exemplary floor plans with particularly pronounced differences between the predicted rent prices of the hedonic model with vs. without floor plans. The floor plans (a) and (b) yielded upward price adjustments, while the floor plans (c) and (d) yielded downward price adjustments.

Fig.~\ref{fig:diff} suggests that incorporating floor plans yields higher rent price predictions in cases in which floor plans convey relatively complex information.
A possible reason is that these floor plans are particularly informative because they convey information that is not clear from the structured data alone.
For instance, floor plan (a) in Fig.~\ref{fig:diff} has an uncluttered entrance, opening up to the entire building and a comparatively high number of windows, which may facilitate higher valuation.
Floor plan (b) in Fig.~\ref{fig:diff} shows a spacious apartment with a large and open-spaced living room, as well as two separated and self-contained private sections, each outfitted with personal bathrooms.
This may be highly price-relevant information that is only available from the apartment floor plan.
On the contrary, floor plan (c) seems to lack features that warrant a higher rent price.
The apartment it is separated into relatively small rooms connected by a very long corridor.
The only point of entrance to a single balcony is through the room farthest away from the entrance.
Likewise, floor plan (d) has only a tiny bathroom and four rooms that are very elongated.
The apartment thus exhibits potentially unfavorable characteristics that are not directly observable from from the structured data.
Altogether, our exploratory analysis suggests that floor plans contain hidden information that is highly price-relevant -- even after accounting for structural and locational characteristics of an apartment.

\section{Discussion}

\textbf{Research implications: }
Our empirical findings contribute to the existing research body by quantifying the hedonic value of floor plans and establishing the link between the visual design of floor plans and real estate prices.
Prior research \cite{Hill.1997,Nowak.2017} suggests that real estate markets tend to contain a variety of non-observable or hidden characteristics that are not taken into account by conventional valuation methods.
Our findings show that there is indeed an underutilization of the available data in current hedonic models.
Specifically, we show that floor plans contain price relevant information in online real estate listings.
This suggests that there are hidden features in floor plans, such as the relative size and positioning between the rooms, that are relevant even after controlling for structural and locational characteristics.
To the best of our knowledge, our paper is the first study to demonstrate that harnessing floor plans can enhance real estate appraisal on online real estate platforms.

\vspace{0.2cm}
\textbf{Practical implications: }
From a practical perspective, our findings are particularly relevant for online real estate platforms.
Currently, the decision of whether or not to upload a floor plan for an apartment is typically left to the discretion of the user.
Based on our finding that floor plans contain price-relevant information that helps users to make an informed decision, real estate platforms should consider making them a mandatory part of the listings and show them more prominently.
This would allow for greater market transparency for both renters and landlords, as well as home buyers and sellers.
In the backend of online real estate platforms, floor plans could also be used as an additional price predictor.
For example, online real estate platforms could implement a model similar to ours after the author has filled all fields to recommend an appropriate rent price.
Alternatively, the proposed prediction could be listed alongside the price set by the author, such that the renter or home buyer has a clearer picture of the value of the property.
Our model could also suggest links to similar lots based not only on the price but on the information from the floor plans, providing a better market overview and diminishing the problem of information asymmetry.
While some valuation projects, such as Zillow, have started taking images into account, floor plans remain severely underutilized.
Ultimately, our study equips real estate investors with the clear benefit of improved risk assessment, allowing them to enhance their portfolios and reduce the possibility of misvaluation, which indirectly improves the quality of financial markets \cite{Hoesli.2015}.

\vspace{0.2cm}
\textbf{Limitations and future research directions: }
Our work provides several avenues for future research that could further enhance the accuracy of recommendation systems on online real estate platforms.
First, it would be interesting to extend our hedonic model by performing textual analysis (\eg, sentiment analysis \cite{Prollochs.2018,Lutz.2019,Lutz.2020}) of apartment descriptions in real estate listings.
Second, while our study was conducted on the rental market of Berlin, the underlying deep learning approach can easily be applied to listings from other marketplaces and locations.
Third, future research could extend our study by studying the hedonic value of different floor plan layouts or other interpretable floor plan features (\eg, windows, doors, relative room sizes).
It is also a promising research direction to study the suitability of floor plans with different characteristics for certain groups of potential users.

\section{Conclusion}

Current hedonic price models in the online real estate market fail to incorporate information about floor plans, as they are typically not part of the structured information of online listings, but rather provided in the form of accompanying images.
For this purpose, this paper investigates to what extent an automated visual analysis of floor plans on online real estate marketplaces can help to enhance real estate appraisal.
We find that the visual design of floor plans has significant explanatory power regarding rent prices -- even after controlling for structured apartment characteristics such as size and location.
Moreover, we demonstrate that harnessing floor plan sentiment results in \SI{10.56}{\percent} more accurate out-of-sample predictions compared to models that only use structural and locational data.
From a practical perspective, our findings provide decision support for real estate investors and have direct implications for online real estate platforms to improve user experience in their real estate listings.

\bibliographystyle{ACM-Reference-Format-no-doi-abbrv}
\balance
\bibliography{main}

\end{document}